\documentclass[
]{ceurart}

\usepackage{listings}
\usepackage{multirow}
\usepackage{tabularx}
\usepackage{array}
\usepackage{makecell}
\usepackage{xurl}
\usepackage[normalem]{ulem} 
\usepackage{alltt}
\usepackage{xcolor}
\usepackage{wrapfig}
\usepackage{fancyvrb}

\begin{document}

\copyrightyear{2026}
\copyrightclause{Copyright for this paper by its authors.
  Use permitted under Creative Commons License Attribution 4.0
  International (CC BY 4.0).}

\conference{Woodstock'22: Symposium on the irreproducible science,
  June 07--11, 2022, Woodstock, NY}

\conference{RAGE-KG 2026: Retrieval-Augmented Generation Enabled by Knowledge Graphs,
  October 25--26, 2026, Bari, Italy}

\title{Enabling Knowledge Graph Understanding at Scale with the EXplore Your Graphs ENgine (EXYGEN)}


\author[1]{Harshdeep Singh}[
orcid=0009-0001-8468-3237,
email=harshdeep.singh@odoma.ch,
]
\address[1]{\href{www.odoma.ch}{Odoma Ltd.}}

\author[1]{Yurui Zhu}[%
orcid=0009-0002-1429-1139,
email=yurui.zhu@odoma.ch,
]

\author[1]{Giovanni Colavizza}[
orcid=0000-0002-9806-084X,
email=giovanni.colavizza@odoma.ch,
url=, 
]
  
\author[1]{Matteo Romanello}[
orcid=0000-0002-7406-6286, 
email=matteo.romanello@odoma.ch,
url=, 
]
\cormark[1]

\cortext[1]{Corresponding author.}

\begin{abstract}
  We present EXYGEN (EXplore Your Graphs ENgine), a framework for knowledge graph understanding that enables conversational access to knowledge graphs (KGs) at scale. We address two questions in sequence.
  
  First, how effectively can LLMs perform text-to-SPARQL generation given only automatically derived structured metadata and small graph samples, rather than task-specific fine-tuning? We integrate VoID descriptions and ShEx schemas into a retrieval-augmented generation (RAG) pipeline and ablate KG-derived context on the SciQA benchmark. Our best configuration---combining ShEx schemas, retrieved triples, and example question-query pairs---reaches an exact match of 0.419 on execution results without any LLM fine-tuning. We further find that lexical metrics such as F1 poorly predict query correctness, and that larger general-purpose LLMs can outperform smaller code-specialized ones once given sufficient context.
  
  Second, having validated the method, we ask how to generate the structured metadata it relies on from very large KGs, where the SPARQL queries required to generate KG metadata become computationally intractable.
  We introduce a predicate-coverage-aware parallel graph sampling strategy that preserves structural diversity while remaining computationally tractable. On OpenCitations Meta and GESIS, it retains high predicate coverage with minimal triple loss and reduces metadata-generation runtime by over 80×; on ORKG, whose fine-grained ontology makes full-graph generation infeasible rather than merely slow, sampling is not just faster but the only tractable path to obtain complete metadata.

Together, these results show that structured schema context and lightweight prompting can substantially reduce reliance on fine-tuning for scalable conversational access to knowledge graphs, though closing the remaining gap to fully fine-tuned approaches will likely require reducing dependence on curated question-query exemplars---whether through synthetic exemplar generation or an iterative, execution-feedback-driven approach---and validating these findings beyond a single benchmark.
\end{abstract}


\maketitle
\section{Introduction}

Over the last few years, Large Language Models (LLMs) and Knowledge Graphs (KGs) have emerged as complementary, mutually reinforcing technologies. LLMs increasingly allow natural language to serve as an interface to semantic KGs, without requiring users to learn structured query languages such as SPARQL \cite{meyer_assessing_2024a}. Conversely, KGs are a valuable source for grounding LLM text generation and reducing hallucinations \cite{agrawal_can_2024, wagner_mitigating_2025}.

This paper addresses two limitations of existing LLM-based KGQA methods, tackled in deliberate order: first validating that a method works, then making it scale.

The first is \textbf{dependence on KG-specific supervision}, characteristic of approaches based on supervised fine-tuning or in-context learning \cite{10.1145/3757923, emonet_llmbased_2024, pan2025firesparqlllmbasedframeworksparql}: such supervision is costly to obtain for each new graph and limits generalization across KG schemas. \citet{mecharnia_performance_2025} illustrate this concretely, fine-tuning Llama-based models for text-to-SPARQL on both DBpedia and Wikidata: their best model reaches a Macro F1 of 60.68\% on DBpedia but only 13.36\% on Wikidata, a gap they attribute to Wikidata's non-human-readable identifiers, which the model cannot infer from pretraining and instead hallucinates. This schema- and identifier-dependence is exactly the generalization limitation we aim to sidestep via KG-derived context rather than parametric supervision, motivating our first question: can off-the-shelf LLMs instead rely on automatically derived, KG-grounded context---endpoint metadata, retrieved triples, and a few example question-query pairs---to generate SPARQL queries without any fine-tuning? Two artifacts recur throughout this KG-derived context: the Vocabulary of Interlinked Datasets (VoID)\footnote{\url{https://www.w3.org/2001/sw/interest/void/}}, an RDF vocabulary for describing a dataset's structural statistics such as class and predicate partitions, and Shape Expressions (ShEx)\footnote{\url{https://shex.io}}, a schema language that expresses these statistics as concise, human-readable shape definitions specifying a class's expected properties, cardinalities, and value types. Together, they give an LLM structured, class-specific constraints on the KG's schema without requiring any task-specific supervision.

The second limitation is the \textbf{scalability} of the metadata generation such context-based methods depend on \cite{emonet_llmbased_2024}. These approaches automatically generate endpoint metadata (e.g., VoID, ShEx) for inclusion in the LLM context, but rely on computation-intensive SPARQL queries---aggregations, Cartesian products---that cause prohibitive computation times and frequent timeouts at scale. Having validated the context-based method, we therefore ask how to produce this metadata efficiently when graph size exceeds current techniques' limits.

These limitations motivate two research questions:
\begin{itemize}
    \item \textbf{RQ1} To what extent can existing LLMs perform text-to-SPARQL generation without task-specific fine-tuning when provided only with automatically extracted, KG-derived context?
    \item \textbf{RQ2} How can the structured metadata (e.g., VoID descriptions and ShEx schemas) on which this method relies be generated efficiently from knowledge graphs whose size exceeds the scalability limits of current techniques?
\end{itemize}

Our key contributions are:
\begin{itemize}
    \item An ablation study of KG-derived context in a Retrieval-Augmented Generation (RAG) pipeline for text-to-SPARQL generation, analyzing the contribution of VoID, ShEx, retrieved triples, and few-shot question-query pairs, and benchmarking several open-weight LLMs on SciQA~\cite{auer_sciqa_2023} to validate the method and provide empirical insights into model behavior under this setting (RQ1).
    \item A predicate-coverage-aware parallel graph sampling strategy that extends structured metadata generation (VoID, ShEx) to scales beyond existing approaches' limits (RQ2).
\end{itemize}

\section{Related work}

This section reviews prior work along the two research questions guiding this paper. Section~\ref{sec:kgqa-related} surveys KGQA approaches, from early semantic parsing pipelines to recent LLM-based methods, situating our fine-tuning-free text-to-SPARQL approach (RQ1). Section~\ref{sec:graph-sampling-related} reviews graph sampling techniques, which inform our predicate-coverage-aware sampling strategy for scaling metadata generation to very large KGs (RQ2).

\subsection{Knowledge Graph Question Answering}
\label{sec:kgqa-related}

Prior to the advent of LLMs, KGQA systems relied primarily on semantic parsing and information retrieval techniques. These approaches translated natural language questions into formal query representations, typically SPARQL, through carefully engineered pipelines involving entity linking, relation detection, and template-based query construction~\cite{berant-etal-2013-semantic, zafar2018formal, sun2020sparqa}. While effective on well-known, encyclopedic knowledge graphs such as Freebase, DBpedia, and Wikidata, these methods struggled to generalize across different KG schemas and required substantial domain-specific engineering for each new graph.

The emergence of LLMs has substantially shifted the KGQA landscape, enabling more flexible and generalizable approaches to generate SPARQL queries from natural language. These approaches fall broadly into two directions. The first fine-tunes LLMs on domain-specific KGs: \cite{rangel2024sparqlgenerationanalysisfinetuning} adapt OpenLLaMA for life science question answering using data augmentation; FIRESPARQL~\cite{pan2025firesparqlllmbasedframeworksparql} achieves strong performance on scholarly KGs such as the ORKG, though structural inconsistencies and semantic inaccuracies in generated queries persist; and \cite{pfeifer_texttosparql_2026} and \cite{vossebeld_learning_2025} train compact models via Group Relative Policy Optimization (GRPO) with execution-based rewards, learning to construct and self-correct SPARQL queries iteratively. The second relies on prompting and context rather than parameter updates: RAG-based methods supply explicitly provided entities, relations, or schema context to reduce ontological errors and improve generation accuracy~\cite{emonet_llmbased_2024, 10.1145/3757923}, while \cite{dobriy_agentic_2026} and \cite{omar_chattykg_2026} extend this with multi-agent, tool-using architectures for federated and conversational KGQA. Our work falls in this second direction.

\subsection{Graph Sampling}
\label{sec:graph-sampling-related}
Graph sampling has been extensively studied as a means of efficiently analyzing large-scale networks while preserving key structural properties. \citet{leskovec2006sampling} provide a comprehensive comparison of sampling strategies and establish a rigorous evaluation framework, assessing methods across node degree distribution, strongly connected component sizes, the hop plot (reachable pairs within $h$ hops), and clustering coefficient distributions, where the clustering coefficient of a node $v$ with $k$ neighbors is defined as the ratio of existing edges to the maximum possible $\frac{k(k-1)}{2}$. Notably, they find that as little as 15\% of the original graph suffices to faithfully represent its macro-level structure.

Random walk-based sampling is among the most widely used approaches for subgraph extraction~\cite{randomwalk2022}, operating by traversing the graph from a seed node within $n$ hops. While computationally tractable and sensitive to local connectivity, these methods can be biased toward densely connected regions. To address this, Markov Chain Monte Carlo methods, particularly the Metropolis-Hastings algorithm~\cite{huebler2008metropolis}, uses an acceptance probability to accept or reject transitions, enabling optimization towards graph samples that preserve key structural properties of the original graph.

More recently, hypergraph-based sampling~\cite{hypergraph2024} extends these ideas to higher-order relational structures, introducing a degree-biased strategy parameterized by $\alpha$, where higher values favor high-degree nodes assumed to contribute disproportionately to structural coherence. This framework applies equally to knowledge graphs, where node degree serves as a proxy for semantic importance.

\section{Data}

This section describes the data underlying our two research questions. Section~\ref{sec:kgqa-benchmark-datasets} introduces the SciQA benchmark used to evaluate the context-based text-to-SPARQL method presented in Section~\ref{sec:kgqa}. Section~\ref{sec:large-scale-knowledge-graphs} then introduces the large-scale knowledge graphs used to test the scalability of the metadata-generation sampling strategy presented in Section~\ref{sec:scaling}.

\subsection{KGQA benchmark datasets}
\label{sec:kgqa-benchmark-datasets}

We use the SciQA dataset~\cite{auer_sciqa_2023} for benchmarking KGQA, which comprises 2,565 natural language question-SPARQL query pairs grounded in the Open Research Knowledge Graph (ORKG), split into training (1,795 pairs), validation (257 pairs), and test (513 pairs) sets. The dataset combines handcrafted and auto-generated question-query pairs, with answers retrieved from the ORKG via a Virtuoso SPARQL endpoint. SciQA has been widely adopted in the KGQA literature~\cite{pan2025firesparqlllmbasedframeworksparql, emonet_llmbased_2024}. We follow this choice given its challenging, multilingual, and domain-specific nature, targeting a KG less likely to appear in LLM pretraining data than the commonly used Wikidata and DBpedia.

\subsection{Large-scale knowledge graphs}
\label{sec:large-scale-knowledge-graphs}

We tested different approaches to sample the graph using three large-scale knowledge graphs of varying sizes: OpenCitations Meta~\cite{Massari_2024}, GESIS~\cite{SDN-10.7802-2969}, and the Open Research Knowledge Graph (ORKG)~\cite{auer_sciqa_2023}, which contain 4.94 billion, 97.52 million, and 1.13 million triples respectively. OpenCitations Meta is a bibliographic metadata database covering publications referenced in the OpenCitations Index, including metadata such as titles, authors, venues, and persistent identifiers. The GESIS Knowledge Graph is an integrated research KG interlinking metadata of scientific resources---such as datasets, publications, and survey instruments---with entities like authors and social science concepts. For the ORKG, the knowledge graph underlying the SciQA benchmark (Section~\ref{sec:kgqa-benchmark-datasets}), we use a frozen dump rather than the live endpoint. Unlike OpenCitations Meta and GESIS, whose bibliographic schemas are comparatively flat, the ORKG encodes a fine-grained ontology of research contributions, comparisons, and scholarly claims, yielding a much larger space of class--predicate combinations relative to its triple count.

\section{Text-to-SPARQL Generation with KG-Derived Contexts}
\label{sec:kgqa}

\begin{figure*}[!ht]
\begin{center}
\includegraphics[width=\textwidth]{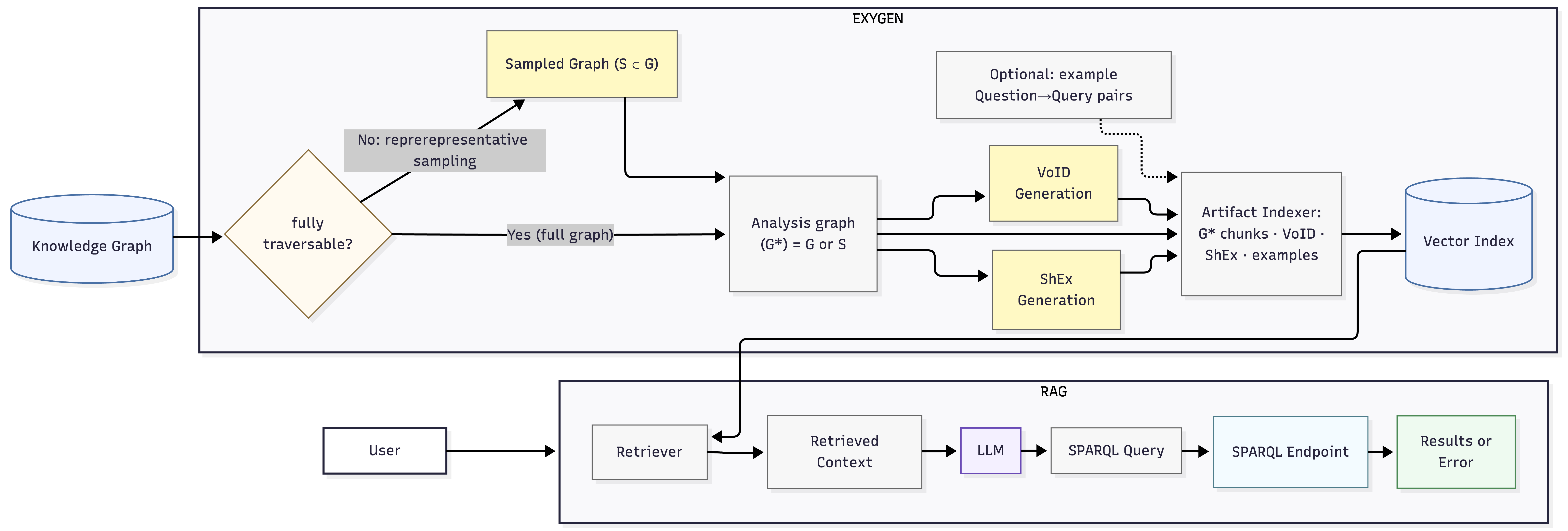}
\caption{The EXYGEN-RAG pipeline. EXYGEN (top) ingests a KG and produces VoID and ShEx artifacts -- via full traversal or representative sampling for large graphs -- indexed for semantic retrieval. The RAG component (bottom) retrieves these artifacts at query time to guide an LLM in generating and executing SPARQL queries against the KG.}
\label{fig:exygen_diagram}
\end{center}
\end{figure*}

In this section we investigate whether off-the-shelf LLMs can generate accurate SPARQL queries with KG-derived context (RQ1). We present EXYGEN, the framework we develop for this purpose, along with the RAG pipeline built on top of it. EXYGEN comprises four components, three of which build the KG-grounded context consumed by our text-to-SPARQL pipeline and are the focus of this section: a VoID description generator that captures dataset-level and per-class/predicate statistics via SPARQL queries; a ShEx extraction pipeline that converts VoID output into structural schema definitions; and a RAG component that retrieves these artifacts at query time to guide an LLM. The fourth component, a predicate-coverage-aware graph sampling module, addresses cases where the KG is too large for metadata generation to run over the full graph; we defer it to Section~\ref{sec:scaling}, as it is only required at scale. The metadata-generation and sampling components are unified under the EXYGEN API\footnote{\url{https://exygen.graphia-ssh.eu/docs}}, a single interface for graph ingestion and schema artifact generation. The downstream RAG pipeline consumes these artifacts but falls outside the EXYGEN API's scope, as shown in Figure~\ref{fig:exygen_diagram}.
 
\subsection{Creation of VoID description}
\label{sec:void-description-creation}
 
The {Vocabulary of Interlinked Datasets} (VoID) is an RDF vocabulary designed to describe linked datasets and their structural metadata. VoID serves as a standardized framework for expressing essential information about RDF datasets, including their structure, content, and interlinking relationships within the broader Linked Data ecosystem. We generate the VoID description using the \texttt{void-generator}\footnote{\url{https://github.com/JervenBolleman/void-generator}} library, which is an open-source project maintained by the Swiss Institute of Bioinformatics. To generate the description, the project uses a series of computationally intensive SPARQL queries (full dataset scans, aggregations) over the endpoint to compute dataset-level statistics and per-class and per-predicate partition metadata.

This challenge is not specific to the \texttt{void-generator} library: \citet{hasnain_sportal_2016} report an analogous pattern in SPORTAL, which issues a comparable battery of self-descriptive SPARQL queries against 618 public endpoints. Query cost scales directly with failure rate: list-only queries (e.g., enumerating class partitions) succeed on up to 94\% of operational endpoints, while aggregation queries computing per-partition instance or triple counts succeed on as few as 25\%, often timing out or returning partial, threshold-capped results (e.g., a fixed cap of 40,000 rows). This confirms that full-scan aggregation cost is inherent to deriving VoID from a live SPARQL interface, motivating both the backend-aware handling below and, for graphs beyond what named-graph selection can mitigate, the sampling approach of Section \ref{sec:scaling}.
 
To support heterogeneous SPARQL endpoints with varying backends (e.g., Virtuoso, QLever), we implement helper functions that identify the backend type via light-weight backend specific queries. Additionally, KGs often contain hundreds of named graphs, making VoID description generation computationally expensive. We extend the existing
\texttt{void-generator} library with a two-stage named graph selection strategy: first, we select the top-$k$ graphs (default $k=10$) by triple count to prioritize knowledge-dense partitions; second, we randomly sample an additional $k$ graphs from the tail of the triple count distribution (i.e., graphs with lower triple counts). This ensures coverage of sparse or disjoint graph partitions, complementing the high-density graphs selected in the first stage.
 
\subsection{Extraction of ShEx schema}
\label{sec:shex-description-creation}

{Shape Expressions} (ShEx) is a schema language for RDF that provides a clean, concise, human-readable syntax for describing and validating the structure of KGs, through shape definitions that specify expected properties, cardinalities, and value types for entities. Unlike VoID, which focuses on descriptive statistics and represents class structures in a convoluted way, ShEx offers a cleaner, more explicit schema representation.

\begin{wrapfigure}[12]{r}{0.42\textwidth}
\begin{Verbatim}
orkgOC:Introduction {
    a [ orkgOC:Introduction ]* ;
    a . + ;
    orkgOP:hasContent xsd:string* ;
    rdfs:label xsd:string*
}
\end{Verbatim}
\captionof{figure}{Example ShEx shape generated for the class \texttt{orkgOC:Introduction}.}
\label{fig:shex-example}
\end{wrapfigure}

We implement a VoID-to-ShEx conversion pipeline that applies regular expressions and SPARQL queries over the VoID description generator's output to identify class definitions and associated property usage, and generate ShEx schemas. As shown in Figure~\ref{fig:shex-example}, instances of \texttt{orkgOC:Introduction} are required to be typed as such (with additional \texttt{rdf:type} values allowed), and each property is constrained by a value type and cardinality derived from the corresponding VoID statistics.

Such schemas provide LLMs with structured, class-specific constraints to improve automated SPARQL query generation.
 
\subsection{RAG-based Question Answering}
\label{sec:rag-question-answering}
 
To generate SPARQL queries from natural-language questions, we use Retrieval-Augmented Generation (RAG)~\cite{gao2024retrievalaugmentedgenerationlargelanguage}. In our setting, RAG retrieves KG-related context via vector similarity and provides the retrieved snippets to the LLM before query generation.
 
We index four types of context, each of which can be used alone or combined:
\begin{itemize}
    \item \textbf{VoID descriptions}: split by double newlines (excluding the prefix-declaration chunk). The remaining chunks contain dataset-level metadata or statistics for specific classes/predicates.
    \item \textbf{ShEx schemas}: split into class-level schema chunks, each describing structural constraints for one class.
    \item \textbf{Retrieved graph triples}: sample RDF triples (provided by KG maintainers or obtained via representative sampling as explained in Section~\ref{sec:scaling}), embedded at the triple level for retrieval.
    \item \textbf{Question-query pairs (QQP)}: example natural language question-SPARQL pairs, provided manually or generated synthetically.
\end{itemize}
 
Each type is maintained in a separate retrieval index, allowing per-type top-$k$ settings that match each type's natural granularity. At inference time, we embed the user question and all indexed chunks using Qwen3-Embedding-0.6B\footnote{\url{https://huggingface.co/Qwen/Qwen3-Embedding-0.6B}}. The vectors are stored in ChromaDB\footnote{\url{https://github.com/chroma-core/chroma}}, and we use LangChain\footnote{\url{https://docs.langchain.com/oss/python/integrations/vectorstores/chroma}} to retrieve the top-$k$ most relevant chunks, which are then appended to the LLM context for SPARQL query generation.
 
\subsection{Experimental Setup}

We evaluate the RAG-based question answering approach described in Section~\ref{sec:rag-question-answering} across multiple configurations to assess the contribution of each contextual component. All experiments are conducted on the SciQA dataset (Section~\ref{sec:kgqa-benchmark-datasets}), using the test set for evaluation and the training set as few-shot context question-query pair examples in the prompt.
 
We conduct experiments across three models of varying parameter scales: the 3B-parameter \texttt{Qwen2.5-Coder-3B-Instruct}\footnote{\url{https://huggingface.co/Qwen/Qwen2.5-Coder-3B-Instruct}} (32k context window), the 30B-parameter (3B active) mixture-of-experts \texttt{Qwen3-Coder-30B-A3B-Instruct}\footnote{\url{https://huggingface.co/unsloth/Qwen3-Coder-30B-A3B-Instruct-GGUF}} (128k context window), and the large-scale \texttt{DeepSeek-V3.1}\footnote{\url{https://huggingface.co/QuantTrio/DeepSeek-V3.1-AWQ}} (128k context window), covering a broad spectrum from lightweight to frontier-scale models. All models were deployed on the Nvidia A100 GPUs with 80GB RAM using the vLLM~\cite{kwon2023efficient} inference library for efficient serving.
 
Each of the three models is evaluated under configurations that progressively combine the available contextual components as mentioned in Section~\ref{sec:rag-question-answering}. 
This systematic ablation allows us to isolate the contribution of each contextual component and assess the benefit of combining them.
We begin with a baseline of no context, then evaluate the effect of individual context types, before testing combinations:
\begin{itemize}
    \item \textbf{Base contexts}: No context, VoID only, ShEx only, retrieved graph triples only, question-query pairs only.
    \item \textbf{Pairwise combinations}: VoID + ShEx, VoID + retrieved graph triples, ShEx + retrieved graph triples, VoID + ShEx+ retrieved graph triples.
    \item \textbf{Few-shot-enhanced}: ShEx + question-query pairs, VoID + ShEx + question-query pairs, VoID + graph triples + question-query pairs, ShEx + graph triples + question-query pairs, and all components combined.
\end{itemize} 

The ``+'' symbol denotes independent per-component retrieval: each context type is queried against its own index with its own top-$k$, and results are concatenated into the LLM context. Components are never pooled into a single index.

For all three models, we use a temperature of 0.1, top-p of 0.9, and a fixed random seed of 42 for reproducibility. Regarding retrieval, for each input question, the 25 most similar chunks are retrieved per contextual component---VoID descriptions, ShEx schemas, and graph triples. For few-shot question-query pair examples, we retrieve the 3 most similar question-query pairs from the SciQA training set based on semantic similarity to the input question, following the approach of~\cite{10.1145/3757923} in selecting the optimal number of example pairs to include in the context.

\paragraph{Evaluation Metrics}

\label{sec:metrics}
We assess each configuration on the SciQA test set by comparing the generated SPARQL query and its execution results against the ground truth. Lexical metrics computed on the query text alone---such as BLEU or ROUGE---can be misleading: small surface differences may drastically change a query's results, while lexically distant queries can be functionally equivalent. Following recent calls for more comprehensive, execution-aware SPARQL evaluation~\cite{taghzouti_t2smetrics_2026}, we report execution-based and query-based metrics jointly.

We report multiple complementary metrics, and highlight the following as the most informative: \textbf{Executability Rate} (does the query run?), \textbf{EM\textsubscript{RelEx}} (do the returned answers match the ground truth?), and \textbf{F1} (query-based surface similarity), to be read with caution, as high F1 does not always imply high accuracy of results.

\textit{Execution-based metrics} compare queries by their behavior when executed against the target SPARQL endpoint:
\begin{itemize}
\item \textbf{Executability Rate}: The proportion of generated queries that are syntactically valid and execute without runtime error. It captures whether a query is runnable, independent of whether its returned results are correct.
\item \textbf{Execution-based Exact Match Relaxed (EM\textsubscript{RelEx})}: We execute both the generated and reference queries and compare the resulting answer sets using the same relaxed-matching criterion applied to queries directly (Levenshtein similarity $\geq$ 0.9; see below). Computing exact match on results rather than query text shifts the criterion from surface-level similarity to functional equivalence, correctly treating lexically distant queries (e.g., differing in filter formulation or property path) as matches when they return the same answers.
\end{itemize}

\textit{Query-based metrics} compare the generated and reference queries at the text level:
\begin{itemize}
\item \textbf{F1 Score}: The harmonic mean of token-level precision and recall (whitespace-split tokenization) between generated and reference queries. We find F1 alone an unreliable indicator of query correctness, as it captures neither executability nor whether the returned results match the ground truth. We therefore also report F1 separately for \textbf{successful queries} (F1\textsubscript{succ}), i.e. generated queries that are syntactically valid and execute without error, and \textbf{failed queries} (F1\textsubscript{fail}), i.e. generated queries that do not execute successfully (e.g., invalid/missing query, syntax error, timeout, runtime execution error).

\item \textbf{Query-based Exact Match Relaxed (EM\textsubscript{RelQ})}: Exact match measures whether the generated query is identical to the reference. Since strict token-by-token comparison penalizes near-perfect queries for superficial differences (e.g., \texttt{C100022} vs.\ \texttt{Model}), we adopt a relaxed variant based on Levenshtein similarity: queries reaching a similarity of at least 0.9, after removing comments and normalizing whitespace, are treated as matches.
\end{itemize}

Beyond these two families, we report \textbf{Prefix Mismatches (PrefMis)}: the proportion of generated queries whose declared namespace prefixes are inconsistent with those actually used (e.g., declaring \texttt{orkgp:} but referencing \texttt{orkgr:}). This diagnostic metric captures a failure invisible to F1's token-overlap scoring, yet one that typically breaks execution entirely---giving it an outsized impact on EM\textsubscript{RelEx} even for queries otherwise nearly identical to the reference.

\begin{table*}[!htbp]
\centering
\small
\setlength{\tabcolsep}{5pt}
\renewcommand{\arraystretch}{1.2}
\newcolumntype{C}{>{\centering\arraybackslash}p{1.05cm}}
\begin{tabular}{l l | C C | C C C C | C}
\hline
& & \multicolumn{2}{c|}{\shortstack{\textbf{Execution-based}\\\textbf{metrics}}} & \multicolumn{4}{c|}{\shortstack{\textbf{Query-based}\\\textbf{metrics}}} & \multicolumn{1}{c}{\shortstack{\textbf{Error}\\\textbf{sources}}} \\
\cline{3-9}
Context & Model & Exec. & EM\textsubscript{RelEx} & F1 & F1\textsubscript{succ} & F1\textsubscript{fail} & EM\textsubscript{RelQ} & PrefMis \\
\hline
\multicolumn{9}{c}{\textbf{Base Contexts}} \\
\hline
\multirow{3}{*}{NoCtx}
& Qwen2.5-3B  & 0.460 & 0.000 & 0.227 & 0.241 & 0.215 & 0.000 & 0.202 \\
& Qwen3-30B   & 0.925 & 0.000 & 0.270 & 0.271 & 0.243 & 0.000 & 0.031 \\
& DeepSeek    & 0.974 & 0.000 & 0.375 & 0.373 & 0.413 & 0.000 & 0.007 \\
\hline
\multirow{3}{*}{VoID}
& Qwen2.5-3B  & 0.602 & 0.000 & 0.346 & 0.368 & 0.311 & 0.000 & 0.220 \\
& Qwen3-30B   & 0.945 & 0.000 & 0.340 & 0.340 & 0.330 & 0.000 & 0.013 \\
& DeepSeek    & 0.947 & 0.000 & 0.372 & 0.372 & 0.369 & 0.000 & \textbf{0.000} \\
\hline
\multirow{3}{*}{ShEx}
& Qwen2.5-3B  & 0.508 & 0.000 & 0.245 & 0.264 & 0.224 & 0.000 & 0.325 \\
& Qwen3-30B   & \underline{0.976} & 0.000 & 0.371 & 0.372 & 0.332 & 0.000 & \underline{0.003} \\
& DeepSeek    & \textbf{0.994} & 0.000 & 0.381 & 0.381 & 0.302 & 0.000 & \textbf{0.000} \\
\hline
\multirow{3}{*}{Retr. triples}
& Qwen2.5-3B  & 0.142 & 0.000 & 0.267 & 0.284 & 0.264 & 0.000 & 0.518 \\
& Qwen3-30B   & 0.943 & 0.000 & 0.339 & 0.339 & 0.337 & 0.000 & 0.011 \\
& DeepSeek    & 0.947 & 0.000 & 0.350 & 0.352 & 0.314 & 0.000 & \textbf{0.000} \\
\hline
\multirow{3}{*}{QQP}
& Qwen2.5-3B  & 0.000 & 0.000 & 0.713 & 0.000 & 0.713 & 0.000 & 0.853 \\
& Qwen3-30B   & 0.000 & 0.000 & \underline{0.758} & 0.000 & 0.758 & 0.000 & 0.884 \\
& DeepSeek    & 0.953 & \textbf{0.097} & \textbf{0.769} & 0.773 & 0.695 & \textbf{0.241} & 0.037 \\
\hline
\multicolumn{9}{c}{\textbf{Pairwise combinations (without question-query pairs)}} \\
\hline
\multirow{3}{*}{VoID+ShEx}
& Qwen2.5-3B  & 0.844 & 0.000 & 0.372 & 0.379 & 0.333 & 0.000 & 0.041 \\
& Qwen3-30B   & 0.881 & 0.000 & 0.295 & 0.302 & 0.237 & 0.000 & 0.038 \\
& DeepSeek    & 0.947 & 0.000 & 0.364 & 0.367 & 0.301 & 0.000 & \textbf{0.000} \\
\hline
\multirow{3}{*}{VoID+Retr. triples}
& Qwen2.5-3B  & 0.499 & 0.000 & 0.339 & 0.350 & 0.328 & 0.000 & 0.155 \\
& Qwen3-30B   & 0.526 & 0.000 & 0.323 & 0.341 & 0.301 & 0.000 & \underline{0.001} \\
& DeepSeek    & 0.830 & 0.000 & 0.307 & 0.318 & 0.250 & 0.000 & \textbf{0.000} \\
\hline
\multirow{3}{*}{ShEx+Retr. triples}
& Qwen2.5-3B  & 0.875 & 0.000 & \underline{0.373} & 0.382 & 0.315 & 0.000 & 0.009 \\
& Qwen3-30B   & 0.984 & 0.000 & 0.346 & 0.346 & 0.296 & 0.000 & 0.003 \\
& DeepSeek    & \underline{0.984} & 0.000 & 0.371 & 0.371 & 0.320 & 0.000 & \textbf{0.000} \\
\hline
\multirow{3}{*}{\makecell[l]{VoID+ShEx\\+Retr. triples}}
& Qwen2.5-3B  & 0.834 & 0.000 & 0.362 & 0.378 & 0.278 & 0.000 & 0.011 \\
& Qwen3-30B   & 0.935 & 0.000 & 0.328 & 0.336 & 0.203 & 0.000 & 0.009 \\
& DeepSeek    & \textbf{0.986} & 0.000 & \textbf{0.379} & 0.379 & 0.341 & 0.000 & \textbf{0.000} \\
\hline
\multicolumn{9}{c}{\textbf{Few-shot-enhanced combinations (with question-query pairs)}} \\
\hline
\multirow{3}{*}{ShEx+QQP}
& Qwen2.5-3B  & 0.001 & 0.000 & \underline{0.886} & 0.461 & 0.887 & 0.000 & 0.982 \\
& Qwen3-30B   & 0.684 & 0.062 & 0.799 & 0.771 & 0.859 & \textbf{0.541} & 0.306 \\
& DeepSeek    & 0.265 & 0.107 & 0.884 & 0.853 & 0.895 & 0.173 & 0.713 \\
\hline
\multirow{3}{*}{VoID+ShEx+QQP}
& Qwen2.5-3B  & 0.001 & 0.000 & 0.876 & 0.461 & 0.877 & 0.000 & 0.980 \\
& Qwen3-30B   & 0.382 & 0.011 & 0.787 & 0.716 & 0.831 & 0.234 & 0.610 \\
& DeepSeek    & \underline{0.980} & \underline{0.329} & 0.760 & 0.759 & 0.774 & 0.053 & \textbf{0.000} \\
\hline
\multirow{3}{*}{\makecell[l]{VoID+Retr. Triples\\+QQP}}
& Qwen2.5-3B  & 0.000 & 0.000 & 0.880 & 0.000 & 0.880 & 0.000 & 0.980 \\
& Qwen3-30B   & 0.001 & 0.000 & \textbf{0.896} & 0.461 & 0.896 & 0.000 & 0.988 \\
& DeepSeek    & 0.000 & 0.000 & 0.779 & 0.000 & 0.779 & 0.000 & 0.869 \\
\hline
\multirow{3}{*}{\makecell[l]{ShEx+Retr. Triples\\+QQP}}
& Qwen2.5-3B  & 0.007 & 0.000 & 0.872 & 0.422 & 0.876 & 0.000 & 0.966 \\
& Qwen3-30B   & 0.844 & 0.037 & 0.735 & 0.728 & 0.773 & \underline{0.483} & 0.144 \\
& DeepSeek    & \textbf{0.986} & 0.272 & 0.765 & 0.764 & 0.790 & 0.273 & \textbf{0.000} \\
\hline
\multirow{3}{*}{\makecell[l]{VoID+ShEx\\+Retr. triples+QQP}}
& Qwen2.5-3B  & 0.003 & 0.000 & 0.872 & 0.459 & 0.873 & 0.000 & 0.966 \\
& Qwen3-30B   & 0.604 & 0.046 & 0.759 & 0.717 & 0.822 & 0.376 & 0.393 \\
& DeepSeek    & 0.978 & \textbf{0.419} & 0.770 & 0.768 & 0.842 & 0.162 & \textbf{0.000} \\
\hline
\end{tabular}
\caption{Text-to-SPARQL results on SciQA across RAG context configurations, grouped by execution-based, query-based, and error-source metrics (defined in Section~\ref{sec:metrics}). For each family of configurations, \textbf{bold} marks the best value and \underline{underline} the second best per column, across all models and contexts therein; F1 is further split into successful vs. failed queries (F1\textsubscript{succ}/F1\textsubscript{fail}).}
\label{tab:kgqa_main_result}
\end{table*}
 
\subsection{Results \& Error Analysis}
\label{sec:kgqa-results-error-analysis}

Table~\ref{tab:kgqa_main_result} contains our full result across different configurations. It shows that, for most base contexts used in isolation, result-level correctness remains near zero: \textit{Exact Match (Relaxed)} is 0 for all models in \textit{NoCtx}, \textit{VoID}, \textit{ShEx}, and \textit{Retrieved Triples}. The main exception is \texttt{DeepSeek-V3.1} with \textit{QQP}, which attains 50 exact matches ($50/513 \approx 0.097$). This contrast also highlights that high executability does not reliably imply correct results: for example, \textit{ShEx} alone yields very high executability for \texttt{Qwen3-Coder-30B-A3B-Instruct-Q8\_0} ($0.9766$) and \texttt{DeepSeek-V3.1-vLLM} ($0.9942$), yet both still obtain zero exact matches in that setting.
This insensitivity extends to F1 itself: across most model--context pairs, F1\textsubscript{succ} and F1\textsubscript{fail} remain close in magnitude, indicating that query-text F1 is largely uninformative about whether a query actually executes. Note that \(F1_{\text{total}}\) can be numerically close to \(F1_{\text{fail}}\) or \(F1_{\text{succ}}\) when one group dominates, since \(F1_{\text{total}}\) is a count-weighted average of the two.

A similar pattern appears for context combinations without QQP. Across all the pairwise settings (\textit{VoID + ShEx}, \textit{ShEx + Retrieved Triples}, \textit{VoID + Retrieved Samples} and \textit{VoID + ShEx + Retrieved Triples}), \textit{Exact Match (Relaxed)} remains 0 for all model--context pairs, despite strong executability in several cases (e.g., \texttt{Qwen3-Coder-30B-A3B-Instruct-Q8\_0} and \texttt{DeepSeek-V3.1-vLLM} reach $\approx 0.98$ in \textit{ShEx + Retrieved Triples}, while \texttt{DeepSeek-V3.1-vLLM} reaches $0.9474$ in \textit{VoID + ShEx}). Thus, improving executability via metadata/schema/triple context alone does not reliably yield correct final answers; example-driven task conditioning (QQP) appears necessary to obtain non-zero exact-match performance in this benchmark.

This is confirmed once pairwise combinations are enriched with QQPs. Here the picture changes sharply, and the gap between model scales widens substantially: \texttt{Qwen2.5-3B} never exceeds an executability of 0.007 in any QQP-enhanced configuration, despite reaching its highest F1 scores in the entire table (0.872--0.886)---a stark illustration that F1 rewards lexical mimicry of the reference query independently of whether the query runs at all, and one that correlates with the near-total prefix inconsistency (PrefMis 0.96--0.98) affecting this model once QQP is introduced. Figure~\ref{fig:prefmis-example} illustrates a representative case of this failure mode, where a near-identical query fails due to an unresolved prefix mismatch.
\texttt{Qwen3-30B} peaks at EM\textsubscript{RelEx} = 0.062 (\texttt{ShEx+QQP}), while \texttt{DeepSeek} reaches 0.419 with \texttt{VoID+ShEx+Retrieved triples+QQP}---the best result in the table by a wide margin, nearly seven times \texttt{Qwen3-30B}'s best. DeepSeek's next-best configurations, \texttt{VoID+ShEx+QQP} (0.329) and \texttt{ShEx+Retrieved triples+QQP} (0.272), all pair ShEx with at least one further context source and register zero PrefMis, indicating that at this scale, richer multi-source schema grounding improves DeepSeek’s ability to produce executable namespace-consistent queries.

ShEx's importance becomes clearest in the one configuration that omits it under QQP: \texttt{VoID+Retrieved triples+QQP} collapses executability to near-zero for all three models alike (0.000, 0.001, 0.000 for \texttt{Qwen2.5-3B}, \texttt{Qwen3-30B}, and \texttt{DeepSeek}, respectively), with PrefMis correspondingly spiking to 0.980, 0.988, and 0.869. This is the clearest evidence in our results that ShEx, rather than model scale, is the decisive factor in whether a generated query is executable at all---its absence is catastrophic regardless of which model is used.

\begin{figure}[!htbp]
\centering
\begin{minipage}[t]{0.48\textwidth}
\centering
\textbf{Ground truth}
\begin{footnotesize}
\begin{alltt}
PREFIX orkgp: <http://orkg.org/orkg/predicate/>
PREFIX orkgc: <http://orkg.org/orkg/class/>
PREFIX orkgr: <http://orkg.org/orkg/resource/>
SELECT DISTINCT ?metric ?metric_lbl
  (MAX(?value) AS ?score)
WHERE \{
  ?dataset a orkgc:Dataset ;
           rdfs:label ?dataset_lbl .
  FILTER (str(?dataset_lbl) =
    "Ball in cup, catch (DMControl100k)")
  ?benchmark orkgp:HAS_DATASET ?dataset ;
    orkgp:HAS_EVALUATION ?eval .
  # ...
\}
\end{alltt}
\end{footnotesize}
\end{minipage}
\hfill
\begin{minipage}[t]{0.48\textwidth}
\centering
\textbf{Generated}
\begin{footnotesize}
\begin{alltt}
PREFIX \textcolor{blue}{orkgOP}: <http://orkg.org/orkg/predicate/>
PREFIX \textcolor{blue}{orkgOC}: <http://orkg.org/orkg/class/>
PREFIX \textcolor{blue}{orkgOR}: <http://orkg.org/orkg/resource/>
SELECT DISTINCT ?metric ?metric_lbl
  (MAX(?value) AS ?score)
WHERE \{
  ?dataset a \textcolor{red}{orkgc}:Dataset ;
           rdfs:label ?dataset_lbl .
  FILTER (str(?dataset_lbl) =
    "Ball in cup, catch (DMControl100k)")
  ?benchmark \textcolor{red}{orkgp}:HAS_DATASET ?dataset ;
    \textcolor{red}{orkgp}:HAS_EVALUATION ?eval .
  # ...
\}
\end{alltt}
\end{footnotesize}
\end{minipage}
\caption{Ground-truth (left) vs.\ generated (right) query for question ``Can you provide the highest benchmark result, including the metric and score, for the Ball in cup, catch (DMControl100k) dataset?''. Despite F1 = 0.9464, the generated query fails on an unresolved prefix mismatch: \textcolor{blue}{declared} longhand prefixes (blue) are never used, while the query body instead references \textcolor{red}{undeclared} shorthand prefixes (red).}
\label{fig:prefmis-example}
\end{figure}

\section{Scaling Metadata Generation to Large Knowledge Graphs}
\label{sec:scaling}

VoID and ShEx generation (Section~\ref{sec:kgqa}) relies on SPARQL queries---aggregations, Cartesian products---that become prohibitively expensive or time out on very large KGs. SPORTAL~\cite{hasnain_sportal_2016} confronts the same bottleneck by tolerating incompleteness, cataloguing whichever statistics each of 618 endpoints can compute within a fixed timeout. This is unsuitable for our setting, where a single target KG must yield complete, consistent metadata for downstream retrieval. We instead reduce the input itself: prior to description generation, we sample the target KG to capture a broad, representative diversity of predicate types, minimizing computational overhead while remaining agnostic to the triple-store backend. We prioritize covering all predicates---the full relational ``vocabulary'' of the graph---to capture its structural diversity rather than only its most frequent patterns.
 
\subsection{Approaches to Representative Graph Sampling}
 
To sample the graph, we perform parallel random walks starting from multiple nodes \textit{N} to ensure diversity in the initial seed set. To traverse the graph, we experimented with three approaches:
 
\begin{itemize}
    \item \textbf{Brute force approach}: At each hop, we consider all neighbors. To mitigate exponential neighborhood expansion, we adopt \textit{pruning}: neighbors are weighted by predicate novelty, with rarer predicates receiving higher weights, and only the top fraction (25\%, 50\%, or 75\%) are retained to reduce redundancy while maintaining coverage.
    \item \textbf{Filtering by high degree}: At each hop, we keep the top-$D$ neighbors by degree (optionally with pruning), under the assumption that high-degree nodes are well-connected and likely link to lower-degree nodes as well. We also test a high+low variant that adds the same number of lowest-degree neighbors, ensuring that sparsely connected or isolated nodes are not systematically excluded from the sample.
    \item \textbf{Filtering by semantic similarity}: At each hop, node embeddings are computed by incorporating both the node's and its neighbors' attributes, retaining only nodes whose similarity scores fall below a set threshold to avoid redundancy. We use the \textit{BAAI/bge-m3 model}\footnote{\url{https://huggingface.co/BAAI/bge-m3}} which supports over 100 languages and has an 8192-token context window, enabling full capture of node attributes without truncation.
\end{itemize}

\subsection{Experimental Setup}
\label{sec:kg-sampling-approaches}

We tested the three approaches to traverse the graph with different parameters on two large-scale knowledge graphs: OpenCitations Meta and GESIS mentioned in Section~\ref{sec:large-scale-knowledge-graphs}.

Table~\ref{tab:representative_sampling} outlines the experimental configurations, defined by the number of start nodes ($N \in \{10, 20, 40\}$), walk hops ($H \in \{1, 2, 3\}$), and pruning percentage ($\alpha \in \{25\%, 50\%, 75\%\}$), together with the neighbor-selection strategy. We use the same set of start nodes across all configurations of a KG to ensure a fair comparison. We evaluate each configuration using four metrics: \textbf{Triples}, the number of triples in the sampled subgraph; \textbf{PredCov(\%)}, the fraction of unique predicates in the full KG observed at least once in the sample; \textbf{MissTrip(\%)}, the fraction of full-KG triples whose predicate is absent from the sample; and \textbf{Time (minutes)}, the sampling runtime.

\begin{table*}[!htbp]
\centering
\footnotesize
\setlength{\tabcolsep}{3pt}
\renewcommand{\arraystretch}{1.12}
\begin{tabularx}{\textwidth}{
    >{\raggedright\arraybackslash}l
    c c c
    >{\raggedright\arraybackslash}X
    r c c c}
\hline
\textbf{Dataset} & \textbf{Nodes} & \textbf{Hops} & \textbf{Prune} & \textbf{Strategy} & \textbf{Triples} & \textbf{PredCov} & \textbf{MissTrip} & \textbf{Time (min)} \\
\hline
 
\multirow[c]{8}{*}{GESIS}
& 10 & 1 & 0.50 & Brute force   & 326      & 0.1808 & 0.2249   & 0.50 \\
& 20 & 2 & 0.50 & Brute force   & 52,681   & 0.3062 & 0.0082   & 1.67 \\
& 20 & 3 & 0.75 & Brute force   & 297,078  & 0.3616 & 0.0098   & 228.10 \\
& 20 & 2 & 0.50 & Top-10 degree & 119,943  & 0.3394 & 0.0105   & 11.92 \\
& 20 & 2 & 0.75 & Top-20 degree & 43,214   & 0.3247 & 0.0126   & 3.36 \\
& 40 & 2 & 0.50 & Top-40 degree & 147,456  & 0.3579 & 0.0071   & 7.90 \\
& 20 & 2 & 0.50 & Sim.\ $<0.6$  & 61,587   & 0.3136 & 0.0146   & 2.10 \\
& 40 & 3 & 0.25 & Sim.\ $<0.8$  & 411,600  & \textbf{0.5786} & \uline{0.0042} & 95.82 \\
\hline
 
\multirow[c]{8}{*}{\makecell[l]{OCM}}
& 10 & 1 & 0.50 & Brute force   & 61       & 0.2368 & 0.5009     & 0.30 \\
& 20 & 2 & 0.50 & Brute force   & 251,302  & \textbf{0.7664} & \uline{0.000049} & 5.02 \\
& 20 & 3 & 0.50 & Brute force   & 303,412  & \textbf{0.7664} & \uline{0.000049} & 90.53 \\
& 20 & 2 & 0.50 & Top-10 degree & 251,302  & \textbf{0.7664} & \uline{0.000049} & 36.78 \\
& 20 & 2 & 0.75 & Top-20 degree & 150,420  & 0.7591 & 0.001690   & 11.81 \\
& 40 & 2 & 0.50 & Top-40 degree & 250,480  & \textbf{0.7664} & \uline{0.000049} & 22.60 \\
& 20 & 2 & 0.50 & Sim.\ $<0.6$  & 100,229  & \textbf{0.7664} & \uline{0.000049} & 2.41 \\
& 20 & 2 & 0.50 & Sim.\ $<0.8$  & 200,335  & \textbf{0.7664} & \uline{0.000049} & 3.42 \\
\hline
\end{tabularx}
\caption{Representative sampling results on GESIS and OpenCitations Meta (OCM). Each row is a configuration of start nodes, hops, pruning ratio, and neighbor-selection strategy. Metrics: sampled \textit{Triples}, predicate coverage (\textit{PredCov}), missing-triple ratio (\textit{MissTrip}), and runtime (\textit{Time}). \emph{Bold} = best PredCov; \uline{underline} = best (lowest) MissTrip.}
\label{tab:representative_sampling}
\end{table*}
 
All experiments were conducted on a machine with 32GB memory. Since our sampling approach relies solely on SPARQL queries issued to an external endpoint, the process is largely I/O-bound and imposes minimal computational requirements on the local machine.

\subsection{Results}
\label{sec:kg-sampling-results}

Table~\ref{tab:representative_sampling} shows the results. For GESIS, semantic similarity (threshold 0.8) achieves the highest predicate coverage (57.86\%) and lowest missing-triple rate (0.42\%), at the cost of substantially higher runtime. Top-40 degree strategy offers competitive coverage (35.79\%) with the lowest missing-triple rate among degree-based approaches (0.71\%) in a moderate 7.90 minutes. Brute force (2 hops, 50\% pruning) is the most efficient configuration overall---30.62\% coverage, 0.82\% missing triples, in just 1.67 minutes---making it preferable when runtime is the priority.

For OpenCitations Meta, multiple configurations reach near-optimal predicate coverage (76.64\%) with minimal missing triples (0.0049\%); semantic similarity ($<0.6$) reaches this in just 2.41 minutes, making it both efficient and effective. Coverage plateaus numerically because most undiscovered predicates are administrative or ontological rather than substantive---e.g., \texttt{sd:resultFormat}, \texttt{dct:modified}, \texttt{rdfs:comment}, \texttt{rdfs:isDefinedBy}, \texttt{owl:versionInfo}---so their absence does not diminish the sample's utility for relationship discovery.

Four conclusions follow. First, parallelizing random walks across multiple start nodes with per-hop pruning substantially cuts runtime without sacrificing sample quality. Second, two-hop walks offer the best trade-off, since further hops expand the neighborhood rapidly for little added coverage. Third, degree-based selection carries non-trivial overhead, as computing node degrees at each hop bottlenecks at scale. Finally, semantic-similarity-based selection improves representativeness within a comparable time budget, at the added cost of embedding inference.

\subsection{Full Graph vs.\ Sampled Metadata Generation}

Having validated the sampling strategy in isolation (Sections~\ref{sec:kg-sampling-approaches}--\ref{sec:kg-sampling-results}), we now put it into practice end-to-end: for each of the three knowledge graphs introduced in Section~\ref{sec:large-scale-knowledge-graphs}, we generate full ShEx and VoID descriptors both directly from the live SPARQL endpoint and from a representative sample, and compare the two processes along runtime and query volume. Table~\ref{tab:metadata-generation} reports this comparison across the three KGs, which vary by three orders of magnitude in size.

The results highlight a stark efficiency gap between querying the full graph via the live endpoint and generating metadata from a local sample. For GESIS KG, sampling reduces generation time from roughly 19 minutes to under 14 seconds---a speedup of over 80$\times$---while covering 0.42\% of the 97M-triple graph. OpenCitations Meta (OCM) shows an even larger gap: 64 minutes on the live endpoint collapses to 11 seconds on a sample covering less than 0.01\% of its 4-billion-triple graph, requiring only 331 of the 604 queries the full graph would demand.

\begin{table}[!htbp]
\centering
\label{tab:metadata-generation}
\begin{tabular}{lrrrrr}
\toprule
\textbf{KG}
  & \textbf{Time (Full)}
  & \textbf{Time (Sampled)}
  & \textbf{Queries (Full)}
  & \textbf{Queries (Sampled)}
  & \textbf{Sampling Rate} \\
\midrule
GESIS KG           & 19 min 21 s  & 13.77 s     & 19,928      & 2,100     & 0.42\%       \\
OCM & 64 min 16 s  & 11.34 s     & 604         & 331       & $<$0.01\%    \\
ORKG               & \textit{N/A} & 43 min 37 s & ${\sim}$18M & 9,800,000 & 18.0\%       \\
\bottomrule
\end{tabular}
\caption{ShEx and VoID generation on live endpoints vs.\ sampled subgraphs
         (5 attempts, concurrency 5).}
\end{table}

ORKG behaves differently, and the reason lies in schema complexity rather than raw graph size. Full generation over ORKG requires an estimated 18 million queries---three orders of magnitude more than OCM, despite ORKG being the smaller graph in absolute triples. Because VoID and ShEx generation issue queries per class, per predicate, and per class--predicate pairing (Section~\ref{sec:kgqa}), query volume scales with the size of the schema vocabulary rather than triple count: ORKG's fine-grained ontology of research contributions and comparisons yields a far larger space of class/predicate combinations than the comparatively flat bibliographic schemas of GESIS and OCM. Representatively covering such a heterogeneous schema requires a correspondingly larger sample (18\% vs.\ $<$1\%), and even so, the sampled run takes 43 minutes. Critically, ORKG is the only case where full-graph generation is not merely slower but infeasible: the live endpoint rate-limits requests before generation can finish, making sampling the sole viable path to a complete descriptor rather than just a faster one.

This contrasts with best-effort strategies such as SPORTAL~\cite{hasnain_sportal_2016}, which sidesteps intractability by tolerating incomplete metadata across a population of endpoints rather than guaranteeing complete coverage for any single one. Our sampling rate is instead calibrated per KG to a predicate-coverage target, letting us trade sample size for runtime while still committing to complete, consistent VoID and ShEx descriptors for the target graph.

\section{Conclusions and future work}
This paper presented EXYGEN, a framework tackling two limitations of LLM-based KGQA: dependence on KG-specific supervision, and the limited scalability of the metadata generation on which fine-tuning-free approaches depend. We validated a text-to-SPARQL method built on KG-grounded context, then scaled its underlying metadata generation to billion-triple graphs.

Executability and correctness prove largely independent: ShEx alone drives execution rates above 0.97 for the larger models, yet no context source or combination without exemplars yields a single correct result (EM\textsubscript{RelEx} = 0 throughout). Few-shot examples are decisive: combined with metadata and retrieved triples, DeepSeek-V3.1 reaches EM\textsubscript{RelEx} = 0.419---nearly seven times the runner-up---though ShEx remains essential, as its removal collapses executability across all models. Of the two prior systems benchmarked on SciQA, only FIRESPARQL reports execution-based scores, and its fine-tuned RelaxedEM(all) = 0.85 sets the gap our fine-tuning-free approach still has to close; \citet{10.1145/3757923} report query-based F1 alone, underscoring how uneven evaluation practice makes such comparisons the exception rather than the norm---exactly the gap a shared framework like t2s-metrics is meant to close. Model scale, not code specialization, matters most once exemplars are introduced: \texttt{Qwen2.5-Coder-3B} attains our highest F1 (up to 0.886) while executing almost no queries, whereas DeepSeek-V3.1 converts context most reliably into correct, executable ones---confirming that lexical metrics cannot substitute for execution-based evaluation, and that larger general-purpose models can outperform smaller code-specialized ones given sufficient context.

For metadata generation at scale, our predicate-coverage-aware sampling cuts generation time over 80$\times$ on GESIS and from 64 minutes to 11 seconds on OpenCitations Meta, covering most predicates with negligible triple loss. ORKG is the exception: its fine-grained ontology demands an estimated 18 million queries despite its comparatively modest size---three orders of magnitude more than the larger but flatter OpenCitations Meta---making full-graph generation infeasible. Schema complexity, not triple count, thus governs tractability, making sampling a real necessity for schema-rich graphs; EXYGEN, supporting several triple-store backends, makes this practical from million- to billion-triple KGs alike.

These results also surface three limitations. Our retrieval of VoID and ShEx shapes relies on semantic similarity over property and class labels, an assumption that does not hold for knowledge graphs with opaque or non-descriptive URIs, such as Wikidata, where retrieval quality is likely to degrade substantially. Relatedly, the dependence on few-shot exemplars noted above means our approach, while fine-tuning-free, is not yet exemplar-free, and its applicability to KGs lacking curated question-query pairs remains open. Finally, our sampling strategy optimizes for predicate coverage and triple loss as proxies for metadata quality; we have not directly measured how sampled versus full metadata affects downstream SPARQL generation accuracy, nor evaluated beyond SciQA and the three KGs studied here, leaving the generality of both contributions to be established.

As future work, we plan to extend EXYGEN along three directions. First, we intend to validate our findings beyond SciQA, testing across additional KGQA benchmarks and knowledge graphs to establish the generality of both the context-based prompting approach and the sampling strategy. Second, we plan to explore an agentic framework that iteratively refines SPARQL queries based on execution feedback on failed or empty-result queries to refine and re-issue them; many failures in Section~\ref{sec:kgqa-results-error-analysis}---prefix mismatches and minor token divergences---could be resolved in a single feedback turn, closing much of the remaining gap to correctness. Third, given the strong dependence on few-shot exemplars identified above, we plan to investigate the feasibility of synthetically generating question-query pairs, which would mitigate reliance on pre-existing data and extend the approach's applicability to KGs where such curated examples do not yet exist.

\begin{acknowledgments}
   This work received funding from the European Union (GRAPHIA, grant ID: 101188018) and from the Swiss State Secretariat for Education, Research and Innovation (SERI).
\end{acknowledgments}

\section*{Declaration on Generative AI}
During the preparation of this work, the author(s) made use of Claude (Sonnet 4.6/Sonnet 5, developed by Anthropic). The tool was used to paraphrase and reword passages, to improve overall writing style, and to check grammar and spelling. All content produced with the assistance of this tool was subsequently reviewed, edited, and verified by the author(s), who take full responsibility for the accuracy, originality, and final content of this publication.

\bibliography{bibliography}

\appendix

\end{document}